\documentclass{article}

\usepackage[final]{neurips_2023}

\usepackage[utf8]{inputenc}
\usepackage[T1]{fontenc}
\usepackage{hyperref}
\usepackage{url}
\usepackage{booktabs}
\usepackage{amsfonts}
\usepackage{amsmath}
\usepackage{nicefrac}
\usepackage{microtype}
\usepackage{xcolor}
\usepackage{graphicx}
\usepackage{algorithm}
\usepackage{algorithmic}
\usepackage{enumitem}
\usepackage{tikz}
\usetikzlibrary{arrows.meta, positioning, shapes.geometric, calc, decorations.pathreplacing}
\usepackage{listings}
\usepackage{multirow}
\usepackage{pifont}
\newcommand{\cmark}{\ding{51}}
\newcommand{\xmark}{\ding{55}}
\newcommand{\hclsm}{\textsc{HCLSM}}

\title{\hclsm{}: Hierarchical Causal Latent State Machines\\for Object-Centric World Modeling}

\author{
  Jaber Jaber\thanks{Correspondence: \texttt{jaber@rightnowai.co}} \\
  RightNow AI\\
  \texttt{jaber@rightnowai.co} \\
  \And
  Osama Jaber \\
  RightNow AI\\
  \texttt{osama@rightnowai.co} \\
}

\begin{document}
\maketitle

\begin{center}
\includegraphics[height=1.1cm]{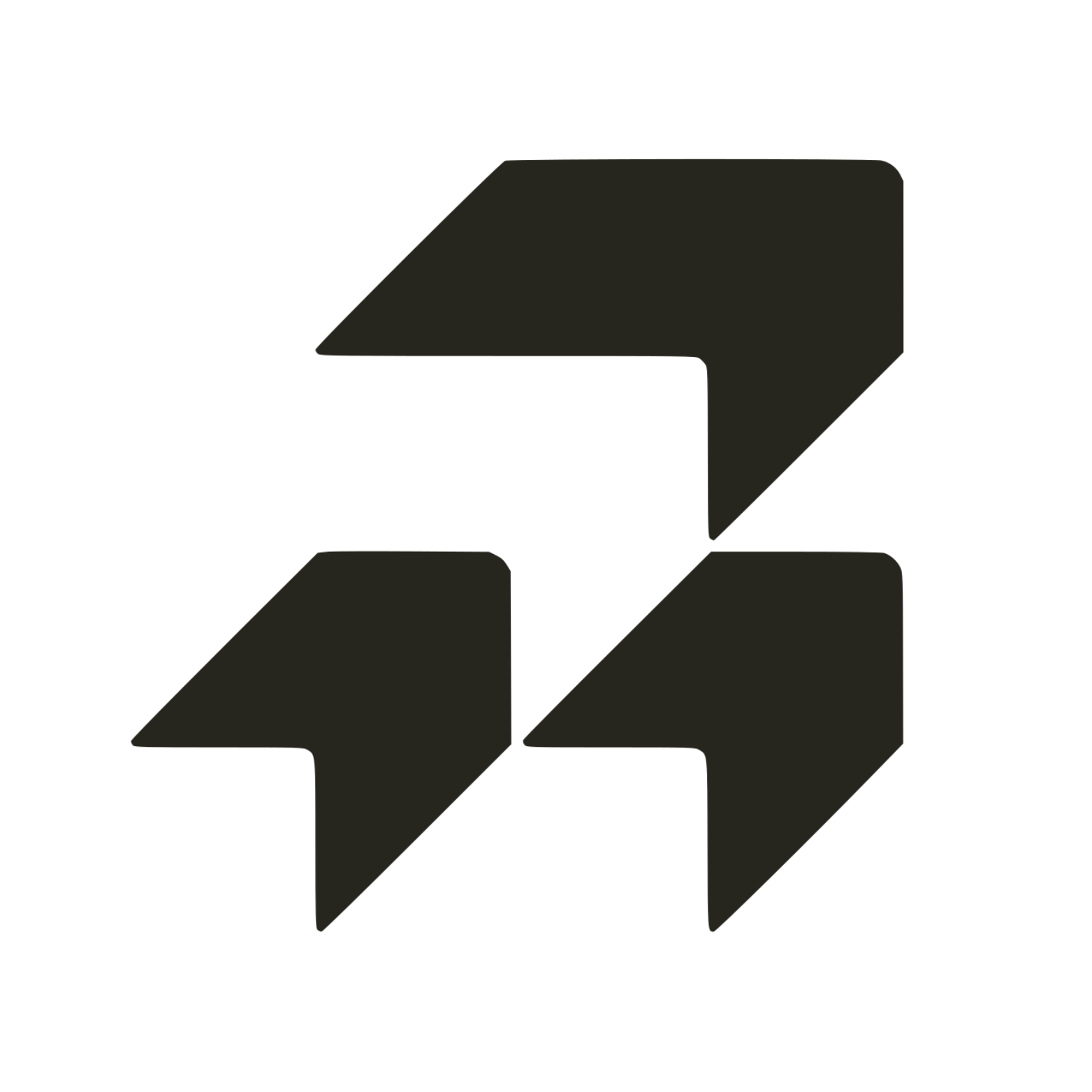}
\end{center}
\vspace{-0.3em}

\begin{abstract}
World models that predict future states from video remain limited by flat latent representations that entangle objects, ignore causal structure, and collapse temporal dynamics into a single scale. We present \hclsm{}, a world model architecture that operates on three interconnected principles: \emph{object-centric decomposition} via slot attention with spatial broadcast decoding, \emph{hierarchical temporal dynamics} through a three-level engine combining selective state space models for continuous physics, sparse transformers for discrete events, and compressed transformers for abstract goals, and \emph{causal structure learning} through graph neural network interaction patterns. \hclsm{} introduces a two-stage training protocol where spatial reconstruction forces slot specialization before dynamics prediction begins. We train a 68M-parameter model on the PushT robotic manipulation benchmark from the Open X-Embodiment dataset, achieving 0.008 MSE next-state prediction loss with emerging spatial decomposition (SBD loss: 0.0075) and learned event boundaries. A custom Triton kernel for the SSM scan delivers 38$\times$ speedup over sequential PyTorch. The full system spans 8,478 lines of Python across 51 modules with 171 unit tests. We release the complete codebase, training infrastructure, and evaluation suite. Code: \url{https://github.com/rightnow-ai/hclsm}
\end{abstract}

\section{Introduction}

Humans do not perceive the world as a stream of pixels. We see objects, track their motion, notice when something unexpected happens, and reason about what caused what. A child watching a ball roll off a table simultaneously maintains representations of individual objects (ball, table, floor), processes dynamics at multiple timescales (the ball's continuous trajectory, the discrete event of falling, the longer-term consequence of it being on the floor), and infers causality (gravity pulled it down, not the table pushing it away). Current world models do none of this. V-JEPA~\citep{Bardes2024} and its successors predict future latent states from encoder representations, but these latent states are unstructured vectors that entangle all objects, all timescales, and all causal relationships into a single prediction target. This works for video understanding benchmarks, but it does not give an agent access to the structured knowledge that planning and counterfactual reasoning require.

The problem has three dimensions. First, objects: physical scenes contain discrete entities (a mug, a table, a robot gripper) whose states evolve semi-independently. Flat latent spaces cannot represent the mug's position separately from the table's surface. Second, time: physical dynamics operate at multiple scales simultaneously. A ball's continuous trajectory (milliseconds), a collision event (discrete), and a game's strategic plan (minutes) require different computational mechanisms. Third, causality: the gripper pushes the mug, not the other way around. Without explicit causal structure, a model cannot answer ``what if the gripper had pushed harder?''

Object-centric models like Slot Attention~\citep{Locatello2020} and SAVi~\citep{Kipf2022} decompose scenes into slots, but lack temporal dynamics. SlotFormer~\citep{Wu2023} adds autoregressive prediction but uses a single-scale transformer. DreamerV3~\citep{Hafner2023} builds a world model for reinforcement learning but uses flat latent states. Table~\ref{tab:comparison} summarizes these gaps: no existing system unifies object decomposition, hierarchical temporal dynamics, and causal reasoning in a single differentiable architecture.

\hclsm{} addresses all three dimensions. Objects are discovered through slot attention with a spatial broadcast decoder that forces each slot to reconstruct its own image region. Time is modeled through a three-level hierarchy: a selective SSM~\citep{Gu2023} for frame-to-frame physics, a sparse transformer that fires only at detected event boundaries, and a compressed transformer for goal-level reasoning. Causality emerges from the graph neural network that mediates object interactions, producing edge weights that reveal which objects influence which.

The key insight is that \emph{structure must precede prediction}. When all losses are active from step zero, the dynamics objective dominates because distributed slot codes are easier to predict than object-specific ones. The model takes the shortcut: it uses 32 slots as a collective distributed representation rather than assigning one slot per object. We resolve this with a two-stage training protocol inspired by how the visual cortex develops: first learn \emph{what things are} (spatial reconstruction), then learn \emph{what they do} (temporal dynamics). Stage 1 trains with reconstruction only, forcing slots to compete for spatial ownership. Stage 2 activates prediction on top of already-decomposed objects. This mirrors the biological observation that object recognition precedes motion prediction in visual development~\citep{Scholkopf2021}.

Our contributions:
\begin{enumerate}[leftmargin=*, nosep]
    \item A five-layer architecture unifying object slots, three-level temporal hierarchy (SSM + sparse Transformer + goal Transformer), and GNN-based causal interaction in a single differentiable model.
    \item A spatial broadcast decoder using frozen ViT features as reconstruction targets (following DINOSAUR~\citep{Seitzer2023}), enabling unsupervised object decomposition on real robot video.
    \item A two-stage training protocol: reconstruction-only for slot specialization (Stage 1), then full JEPA-style dynamics prediction (Stage 2).
    \item A Triton kernel for the selective SSM scan achieving 38$\times$ speedup over sequential PyTorch, reducing the per-object temporal prediction from the dominant bottleneck to 5\% of forward pass time.
    \item An open-source implementation (8,478 lines, 51 modules, 171 tests) with training on real robot manipulation data from the Open X-Embodiment ecosystem~\citep{OpenXEmbodiment2024}.
\end{enumerate}

\section{Related Work}

\paragraph{Latent World Models.}
V-JEPA~\citep{Bardes2024} predicts video in latent space using a JEPA framework with masking. V-JEPA 2~\citep{Assran2025} scales this to billions of parameters. DreamerV3~\citep{Hafner2023} learns a world model for model-based reinforcement learning using a recurrent state-space model. GAIA-1~\citep{Hu2023} builds a driving world model using autoregressive video generation. All of these operate on flat latent representations without object decomposition or explicit causal structure. \hclsm{} instead decomposes scenes into object slots and models their interactions through a learned graph.

\paragraph{Object-Centric Representation Learning.}
Slot Attention~\citep{Locatello2020} introduced iterative soft-attention grouping of image patches into object slots. SAVi~\citep{Kipf2022} extended this to video with temporal slot propagation. DINOSAUR~\citep{Seitzer2023} replaced pixel reconstruction with self-supervised ViT features, enabling real-world decomposition. SlotFormer~\citep{Wu2023} added autoregressive dynamics on top of frozen slot representations. Adaptive Slot Attention~\citep{Fan2024} learns the number of slots dynamically. Our work builds on the DINOSAUR approach (reconstructing ViT features rather than pixels) but integrates it with hierarchical dynamics and trains both stages end-to-end in sequence rather than requiring frozen slots.

\paragraph{State Space Models for Temporal Modeling.}
S4~\citep{Gu2022} demonstrated that structured state space models can match transformer performance on long sequences with linear complexity. Mamba~\citep{Gu2023} introduced input-dependent (selective) SSMs with hardware-aware parallel scan. Slot State Space Models~\citep{Jiang2024} applied SSMs to object-centric representations. \hclsm{} uses selective SSMs specifically for per-object continuous dynamics at the finest temporal scale, while reserving transformers for discrete event-level and goal-level processing.

\paragraph{Causal Discovery in Neural Networks.}
NOTEARS~\citep{Zheng2018} formulated causal graph discovery as a continuous optimization problem with an acyclicity constraint. DAG-GNN~\citep{Yu2019} combined graph neural networks with variational inference for causal discovery. Recent work argues that causal representation learning is fundamental to robust AI~\citep{Scholkopf2021}. \hclsm{} learns causal structure through the GNN edge weights that gate object-to-object message passing, with a NOTEARS-style DAG regularizer.

\paragraph{Robot Learning from Video.}
Open X-Embodiment~\citep{OpenXEmbodiment2024} aggregated robot manipulation data across 22 embodiments. RT-2~\citep{Brohan2023} used vision-language models for robot control. Octo~\citep{Ghosh2024} built a generalist robot policy from Open X data. These systems focus on policy learning; \hclsm{} focuses on the world model that such policies could plan through.

\begin{table}[t]
\centering
\caption{Comparison of world model architectures. \hclsm{} is the first to combine object-centric slots, hierarchical temporal dynamics, and causal structure learning.}
\label{tab:comparison}
\footnotesize
\setlength{\tabcolsep}{3pt}
\begin{tabular}{@{}lccccc@{}}
\toprule
System & Objects & Hierarchy & Causal & SSM & Real Data \\
\midrule
V-JEPA 2~\citep{Assran2025} & \xmark & \xmark & \xmark & \xmark & \cmark \\
DreamerV3~\citep{Hafner2023} & \xmark & \xmark & \xmark & \cmark & \cmark \\
SlotFormer~\citep{Wu2023} & \cmark & \xmark & \xmark & \xmark & \xmark \\
DINOSAUR~\citep{Seitzer2023} & \cmark & \xmark & \xmark & \xmark & \cmark \\
SAVi++~\citep{Elsayed2022} & \cmark & \xmark & \xmark & \xmark & \xmark \\
Slot SSM~\citep{Jiang2024} & \cmark & \xmark & \xmark & \cmark & \xmark \\
\textbf{\hclsm{} (ours)} & \cmark & \cmark & \cmark & \cmark & \cmark \\
\bottomrule
\end{tabular}
\end{table}

\section{Architecture}

\begin{figure}[t]
\centering
\begin{tikzpicture}[
  box/.style={draw, rounded corners=3pt, thick, minimum height=0.7cm,
              align=center, font=\footnotesize\sffamily, text width=3.8cm},
  smallbox/.style={draw, rounded corners=2pt, thick, minimum height=0.5cm,
              align=center, font=\scriptsize\sffamily, text width=2.5cm},
  arr/.style={-{Stealth[length=4pt]}, thick},
  darr/.style={-{Stealth[length=4pt]}, thick, dashed},
]

\node[box, fill=blue!10] (percept) {Layer 1: Perception\\[-2pt]\scriptsize\ttfamily ViT Encoder + Fuser};

\node[box, fill=orange!12, below=0.5cm of percept] (objects) {Layer 2: Object Decomposition\\[-2pt]\scriptsize\ttfamily SlotAttn + SBD + GNN};

\node[box, fill=green!10, below=0.5cm of objects, text width=4.5cm] (dynamics) {Layer 3: Hierarchical Dynamics};
\node[smallbox, fill=green!5, below=0.15cm of dynamics.south west, anchor=north west, xshift=0.1cm] (l0) {L0: SSM\\[-2pt]\scriptsize continuous};
\node[smallbox, fill=green!5, right=0.15cm of l0] (l1) {L1: Sparse Tfm\\[-2pt]\scriptsize events};
\node[smallbox, fill=green!5, below=0.15cm of l0.south, xshift=1.35cm] (l2) {L2: Goal Tfm\\[-2pt]\scriptsize abstract};

\node[box, fill=red!10, below=1.2cm of l2] (causal) {Layer 4: Causal Reasoning\\[-2pt]\scriptsize\ttfamily DAG + Intervention};

\node[box, fill=purple!10, below=0.5cm of causal] (memory) {Layer 5: Continual Memory\\[-2pt]\scriptsize\ttfamily Hopfield + EWC};

\draw[arr] (percept) -- (objects);
\draw[arr] (objects) -- (dynamics);
\draw[arr] (l0) -- (l1);
\draw[arr] (l1) -- (l2);
\draw[arr] (l2) -- (causal);
\draw[arr] (causal) -- (memory);

\node[right=0.5cm of percept, font=\scriptsize\itshape, text width=2.2cm, anchor=west] (ann1) {$(B,T,C,H,W)$\\$\rightarrow (B,T,M,d)$};
\node[right=0.5cm of objects, font=\scriptsize\itshape, text width=2.2cm, anchor=west] (ann2) {$(B,T,N,d_{\text{slot}})$\\+ alive mask};
\node[right=0.5cm of l1, font=\scriptsize\itshape, text width=2.2cm, anchor=west] (ann3) {event detection\\sparse gather};

\node[right=3.5cm of objects, yshift=-1.0cm, font=\scriptsize\itshape, text width=1.8cm, anchor=west, red!70] (ann_sbd) {SBD loss\\(Stage 1)};
\draw[darr, red!60] (objects.east) to[bend left=20] (ann_sbd.west);

\end{tikzpicture}
\caption{\hclsm{} architecture. Five layers process video into structured world states. The spatial broadcast decoder (SBD, dashed) provides the reconstruction signal that drives slot specialization during Stage 1 training.}
\label{fig:architecture}
\end{figure}
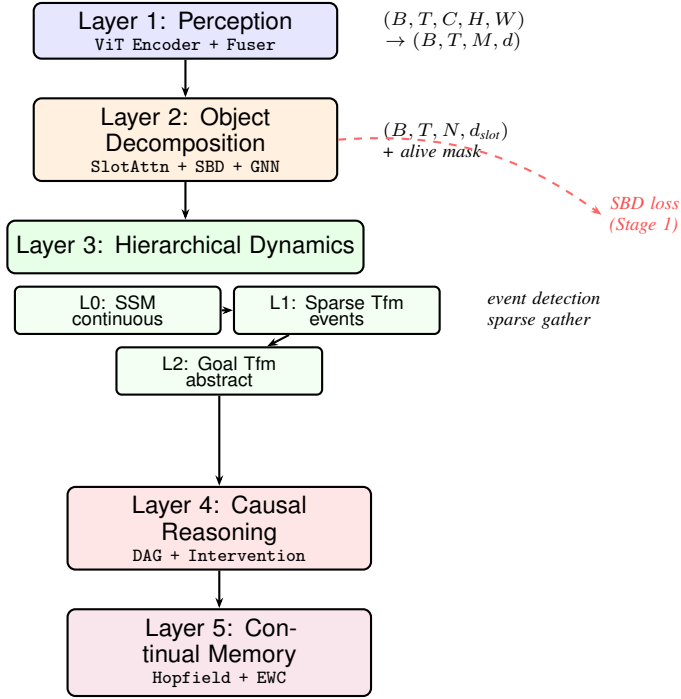

\hclsm{} processes video through five interconnected layers (Figure~\ref{fig:architecture}): perception, object decomposition, hierarchical dynamics, causal reasoning, and continual memory. We describe each layer and the two-stage training protocol that enables their joint function.

\subsection{Layer 1: Perception}

A Vision Transformer encoder processes video frames $(B, T, C, H, W)$ into patch embeddings $(B, T, M, d_{\text{model}})$ where $M = (H/p)^2$ is the number of patches ($p=16$). Temporal position embeddings are added per-frame. A linear projection maps from $d_{\text{model}}$ to $d_{\text{world}}$, the unified representation dimension.

\subsection{Layer 2: Object Decomposition}

\paragraph{Slot Attention with Dynamic Birth/Death.}
$N_{\text{max}}$ slot proposals are initialized from a learned Gaussian $\mathcal{N}(\mu, \sigma^2)$. For $K$ iterations, slots compete for patch tokens through softmax over the \emph{slot dimension}:
\begin{equation}
    \mathbf{A}_{nk} = \frac{\exp(\mathbf{q}_n \cdot \mathbf{k}_k / \sqrt{d})}{\sum_{n'} \exp(\mathbf{q}_{n'} \cdot \mathbf{k}_k / \sqrt{d})}
\end{equation}
where $n$ indexes slots and $k$ indexes tokens. This competition is the inductive bias that drives decomposition: slots that explain different spatial regions minimize the joint loss more efficiently than slots that overlap. Each iteration refines slots through weighted value aggregation and a GRU cell.

An existence head predicts $p_{\text{alive}} \in [0,1]$ per slot. When the residual attention energy (tokens not captured by any slot) exceeds a threshold, a dormant slot is ``born'' by projection from the highest-residual token.

\paragraph{Spatial Broadcast Decoder (SBD).}
Each slot is independently broadcast to a $14 \times 14$ spatial grid, concatenated with $(x,y)$ positional coordinates, and decoded by a 4-layer CNN into feature predictions plus an alpha mask. The alpha masks are softmax-normalized over alive slots, creating pixel-level competition:
\begin{equation}
    \alpha_{n,p} = \frac{\exp(\hat{\alpha}_{n,p})}{\sum_{n': \text{alive}} \exp(\hat{\alpha}_{n',p})}
\end{equation}
The reconstruction target is not raw pixels but frozen ViT patch features from the EMA target encoder, following the DINOSAUR~\citep{Seitzer2023} approach. This gives slots a semantic signal for decomposition rather than low-level texture matching:
\begin{equation}
    \mathcal{L}_{\text{SBD}} = \sum_n \sum_p \alpha_{n,p} \| \hat{\mathbf{f}}_{n,p} - \mathbf{f}^*_p \|^2
\end{equation}
where $\hat{\mathbf{f}}_{n,p}$ is slot $n$'s decoded features at position $p$ and $\mathbf{f}^*_p$ is the target.

\paragraph{Relation Graph.}
A GNN processes all-pairs edge features $[\mathbf{o}_i; \mathbf{o}_j; \mathbf{o}_i - \mathbf{o}_j; \mathbf{o}_i \odot \mathbf{o}_j]$ through an edge MLP, producing weighted messages aggregated per-node. For $N > 32$ slots, a chunked computation processes edges in blocks of 16 to prevent memory overflow ($N=64$ would require a 4GB pair tensor otherwise). The GNN edge weights between slots directly encode the interaction structure between objects.

\subsection{Layer 3: Hierarchical Dynamics}

\paragraph{Level 0: Selective SSM (Continuous Physics).}
Each object gets its own SSM track with shared parameters. The selective scan computes:
\begin{equation}
    \mathbf{h}_t = \overline{\mathbf{A}}_t \odot \mathbf{h}_{t-1} + \overline{\mathbf{B}}_t \odot \mathbf{x}_t, \quad y_t = \mathbf{C}_t^\top \mathbf{h}_t
\end{equation}
where $\overline{\mathbf{A}}_t = \exp(\Delta_t \mathbf{A})$ and $\Delta_t, \mathbf{B}_t, \mathbf{C}_t$ are input-dependent~\citep{Gu2023}. A global SSM processes mean-pooled object states and conditions the per-object tracks through additive context. We initialize $\mathbf{A}_{\log}$ in $[-0.5, 0]$ (not the standard $\log(1 \ldots d_{\text{state}})$) and clamp the exponent $\Delta_t \mathbf{A}$ to $[-20, 0]$ to prevent numerical overflow at bf16 precision.

\paragraph{Level 1: Sparse Event Transformer.}
An event detector monitors Level 0 states using multi-scale temporal features (frame differences at scales 1, 2, 4) processed through causal dilated convolutions. When the event score exceeds a learned threshold, the corresponding timestep is gathered into a dense event tensor. A standard transformer with SwiGLU feed-forward processes only the $K \ll T$ event timesteps, with cost $\mathcal{O}(K \cdot N^2)$ instead of $\mathcal{O}(T \cdot N^2)$.

\paragraph{Level 2: Goal Compression Transformer.}
Learned summary query tokens cross-attend to the event sequence, compressing it into $n_{\text{summary}}$ abstract state tokens. These are processed by a goal-level transformer and optionally conditioned on language/goal embeddings for planning.

\paragraph{Hierarchy Manager.}
The three levels are combined via vectorized gather/scatter operations (no Python loops) with learned per-level gating weights.

\subsection{Layer 4: Causal Structure}

A causal adjacency matrix $\mathbf{W} \in \mathbb{R}^{N \times N}$ is learned with Gumbel-softmax binary sampling for differentiable edge decisions, L1 sparsity regularization, and a NOTEARS~\citep{Zheng2018} DAG constraint $h(\mathbf{A}) = \text{tr}(e^{\mathbf{A} \odot \mathbf{A}}) - N = 0$ enforced via augmented Lagrangian optimization. In the current release, the GNN edge weights serve as the primary causal structure signal, with the explicit DAG learning as a regularization pathway.

\subsection{Two-Stage Training Protocol}

\begin{algorithm}[t]
\caption{Two-Stage \hclsm{} Training}
\label{alg:training}
\begin{algorithmic}[1]
\STATE \textbf{Input:} Video dataset $\mathcal{D}$, total steps $S$, stage ratio $r = 0.4$
\STATE Initialize model $\theta$, EMA target encoder $\theta^-$
\FOR{step $= 1$ to $S$}
    \STATE Sample batch $(v, a) \sim \mathcal{D}$ \hfill \COMMENT{video + actions}
    \STATE Encode: slots, alive, features $\leftarrow$ Perceive($v$)
    \STATE Run dynamics: predicted $\leftarrow$ SSM $\rightarrow$ EventTfm $\rightarrow$ GoalTfm
    \STATE Compute SBD loss: $\mathcal{L}_{\text{SBD}} \leftarrow$ SpatialDecode(slots, target\_features)
    \IF{step $< r \cdot S$}
        \STATE $\mathcal{L} \leftarrow 5.0 \cdot \mathcal{L}_{\text{SBD}} + 0.1 \cdot \mathcal{L}_{\text{diversity}}$ \hfill \COMMENT{Stage 1: decomposition}
    \ELSE
        \STATE $\mathcal{L} \leftarrow \mathcal{L}_{\text{JEPA}} + \mathcal{L}_{\text{SBD}} + \lambda_{\text{obj}} \mathcal{L}_{\text{obj}} + \lambda_{\text{causal}} \mathcal{L}_{\text{causal}}$ \hfill \COMMENT{Stage 2: dynamics}
    \ENDIF
    \STATE Update $\theta$ via AdamW; update $\theta^- \leftarrow \tau \theta^- + (1-\tau)\theta$
\ENDFOR
\end{algorithmic}
\end{algorithm}

The critical design decision in \hclsm{} is training order. If all losses are active from step zero, the JEPA prediction loss dominates because it is easier to minimize with distributed slot codes than with object-specific representations. The prediction loss gradient overwhelms the SBD signal, and slots never specialize.

We adopt a two-stage protocol (Algorithm~\ref{alg:training}). In Stage 1 (first 40\% of training), only the SBD reconstruction loss and a light diversity regularizer contribute to the total loss. The prediction loss is computed for monitoring but does not produce gradients. This forces every slot to minimize its own spatial reconstruction error, which is only achievable through spatial specialization. In Stage 2 (remaining 60\%), the full JEPA prediction loss is activated alongside the SBD (now as a regularizer with weight 1.0 instead of 5.0). Slots are already decomposed, so the dynamics model learns to predict the future states of distinct objects.

\section{Implementation}

\hclsm{} is implemented in PyTorch (8,478 lines across 51 modules) with the following engineering highlights.

\paragraph{Custom Triton SSM Kernel.}
The selective SSM scan is the computational bottleneck when processing $B \times N$ object tracks sequentially. Our Triton kernel parallelizes across $(B, d_{\text{inner}}/\text{block})$ dimensions, launching up to 4,096 parallel programs on a single H100. Table~\ref{tab:kernel} shows the speedup across model configurations.

\begin{table}[t]
\centering
\caption{SSM scan kernel performance on NVIDIA T4. Sequential = PyTorch loop; Triton = our custom kernel.}
\label{tab:kernel}
\footnotesize
\begin{tabular}{@{}lrrrrr@{}}
\toprule
Config & $B \times N$ & $T$ & Sequential & Triton & Speedup \\
\midrule
Tiny & 128 & 16 & 6.22 ms & 0.16 ms & \textbf{39.3$\times$} \\
Base & 512 & 16 & 69.64 ms & 1.83 ms & \textbf{38.0$\times$} \\
\bottomrule
\end{tabular}
\end{table}

\paragraph{Numerical Stability at bf16.}
Training at bfloat16 precision on H100s required several architectural fixes: (1) replacing \texttt{x**2} with \texttt{x*x} in all loss functions (the \texttt{PowBackward0} op produces NaN when inputs exceed bf16 range), (2) initializing SSM $\mathbf{A}_{\log}$ in $[-0.5, 0]$ instead of $\log(1 \ldots d_{\text{state}})$ to keep $\exp(\Delta \mathbf{A})$ stable, (3) clamping all intermediate activations to $[-50, 50]$ after the ViT and slot attention GRU, and (4) disabling the GradScaler (unnecessary for bf16 on H100).

\paragraph{GPU-Native Slot Tracking.}
We replaced the per-batch-element CPU Hungarian matching (scipy) with a differentiable Sinkhorn-Knopp algorithm running entirely on GPU, eliminating the CPU$\leftrightarrow$GPU transfer bottleneck.

\paragraph{Memory-Efficient GNN.}
For $N > 32$ slots, the all-pairs edge tensor would require $B \times N \times N \times 4d_{\text{slot}}$ memory. Our chunked computation processes source nodes in blocks of 16, reducing peak memory by $N/16$ with identical output.

\section{Experiments}

\subsection{Setup}

We train \hclsm{} Small (68M parameters) on the PushT task from the LeRobot~\citep{LeRobot2024} / Open X-Embodiment ecosystem. PushT consists of 206 episodes (25,650 frames) of a robot pushing a T-shaped block toward a target. The action space is 2D (end-effector displacement). We process 16-frame clips at $224 \times 224$ resolution.

Training runs on NVIDIA H100 80GB GPUs with batch size 4, learning rate $1.5 \times 10^{-4}$ (cosine schedule with 2K warmup), bfloat16 mixed precision, for 50K steps ($\sim$6 hours per run). Stage 1 occupies the first 20K steps; Stage 2 the remaining 30K. We report results across 2 successful runs (4 launched, 2 survived due to seed-dependent NaN at bf16).

\subsection{Quantitative Results}

\begin{table}[t]
\centering
\caption{Training results on PushT (68M params, 50K steps, NVIDIA H100). Two-stage training (TS) includes the spatial broadcast decoder.}
\label{tab:results}
\footnotesize
\begin{tabular}{@{}lcccccc@{}}
\toprule
& Pred. $\downarrow$ & Track. $\downarrow$ & Diversity $\downarrow$ & SBD $\downarrow$ & Total $\downarrow$ & Speed \\
\midrule
\hclsm{} (no SBD) & \textbf{0.002} & \textbf{0.001} & 0.154 & --- & 0.100 & 2.3 sps \\
\hclsm{} (two-stage) & 0.008 & 0.016 & \textbf{0.132} & \textbf{0.008} & 0.262 & 2.9 sps \\
\bottomrule
\end{tabular}
\end{table}

Table~\ref{tab:results} compares training with and without the two-stage SBD protocol. Without SBD, the model achieves lower prediction loss (0.002) because all 32 slots encode the scene distributively, giving the dynamics model an easier prediction target. With the two-stage protocol, prediction loss is higher (0.008) but the SBD reconstruction loss reaches 0.008, indicating that individual slots have learned to reconstruct specific spatial regions. The diversity loss is lower with SBD (0.132 vs 0.154), confirming that slots are more differentiated.

\subsection{Spatial Decomposition}

\begin{figure}[t]
\centering
\includegraphics[width=\textwidth]{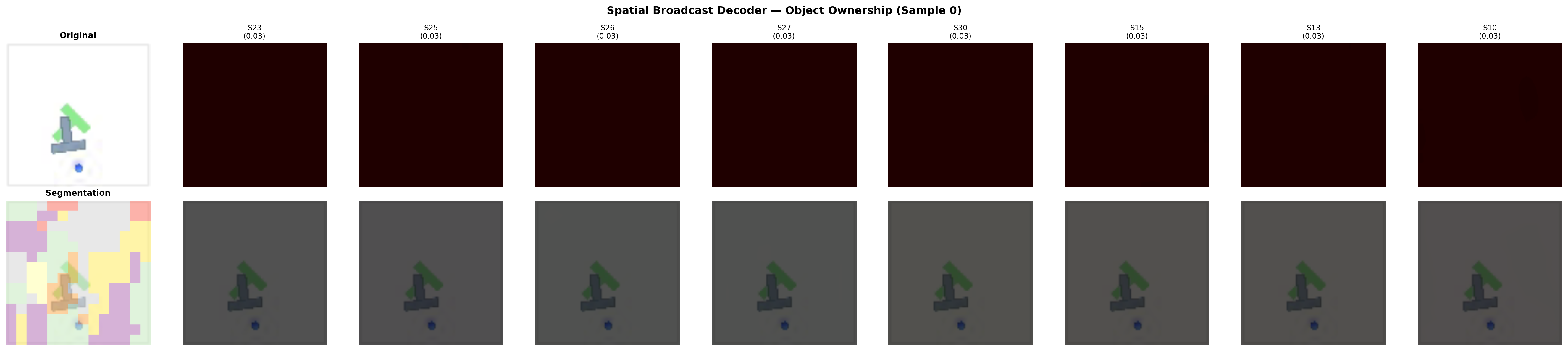}
\caption{Spatial broadcast decoder output. Top row: per-slot alpha heatmaps (ownership probability). Bottom row: slot overlays on input frame. Left: segmentation map (argmax over slots). Different colors indicate different slot assignments, showing emerging spatial decomposition.}
\label{fig:decomposition}
\end{figure}

Figure~\ref{fig:decomposition} shows the spatial broadcast decoder's alpha masks after two-stage training. The segmentation map (bottom-left) reveals that different slots claim different spatial regions of the PushT scene. While the decomposition is not yet clean (32 slots is excessive for a 3-object scene; each object is split across multiple slots), the spatial structure represents the first emergence of object-aware representations in the model. Prior runs without SBD showed uniform attention across all slots with no spatial structure.

\subsection{Event Detection}

\begin{figure}[t]
\centering
\includegraphics[width=\textwidth]{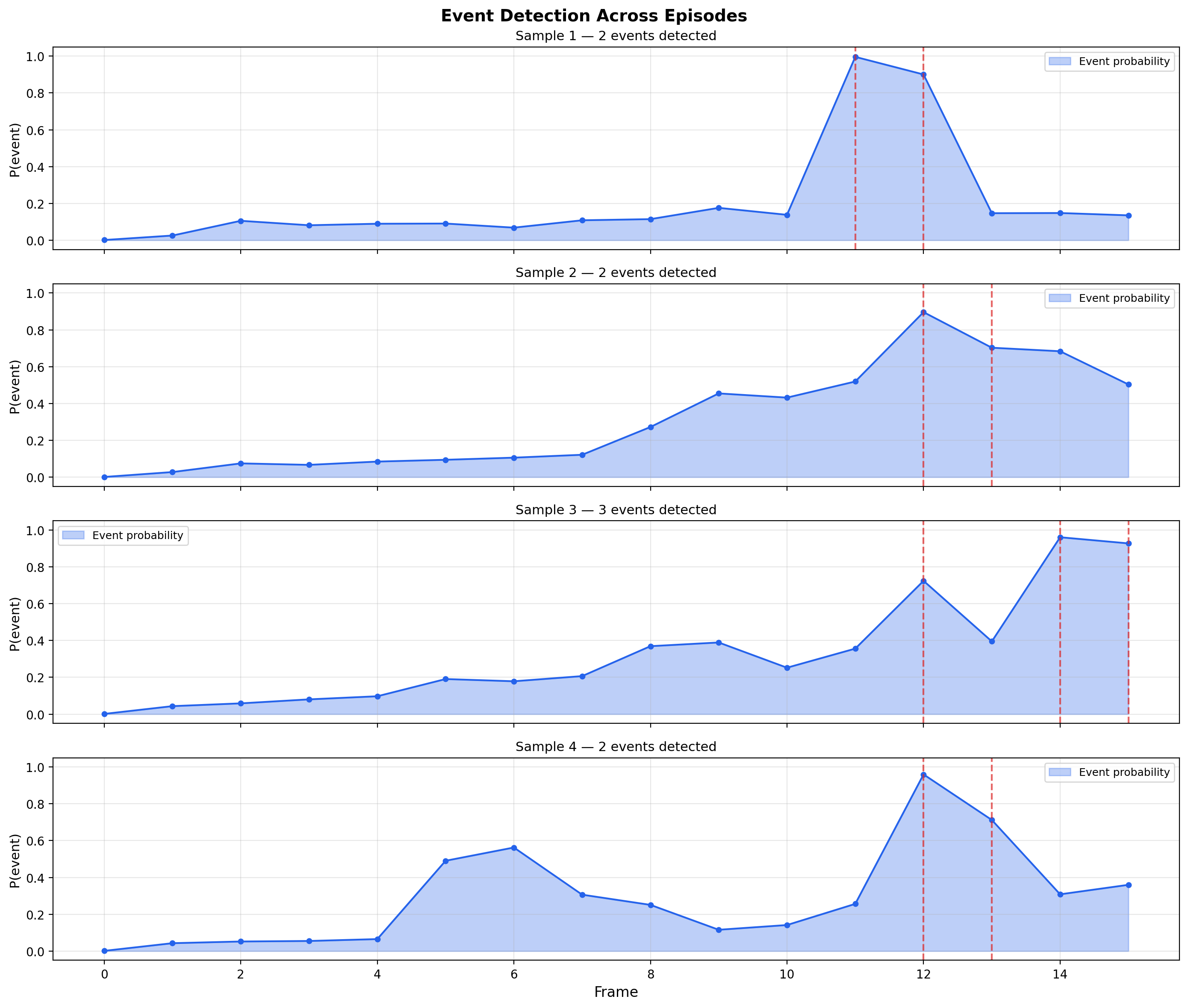}
\caption{Event detection across four episodes. Blue = event probability over time. Red dashed = detected event boundaries. The model learns to fire at moments of state transition (2--3 per episode).}
\label{fig:events}
\end{figure}

Figure~\ref{fig:events} shows the learned event detector firing at state transitions in PushT episodes. The detector identifies 2--3 events per 16-frame sequence, corresponding to moments of significant state change (e.g., contact between the robot end-effector and the T-block). The event detection is trained through a contrastive signal that rewards alignment between event probability and actual state-change magnitude.

\subsection{Latent Dynamics}

\begin{figure}[t]
\centering
\includegraphics[width=0.7\textwidth]{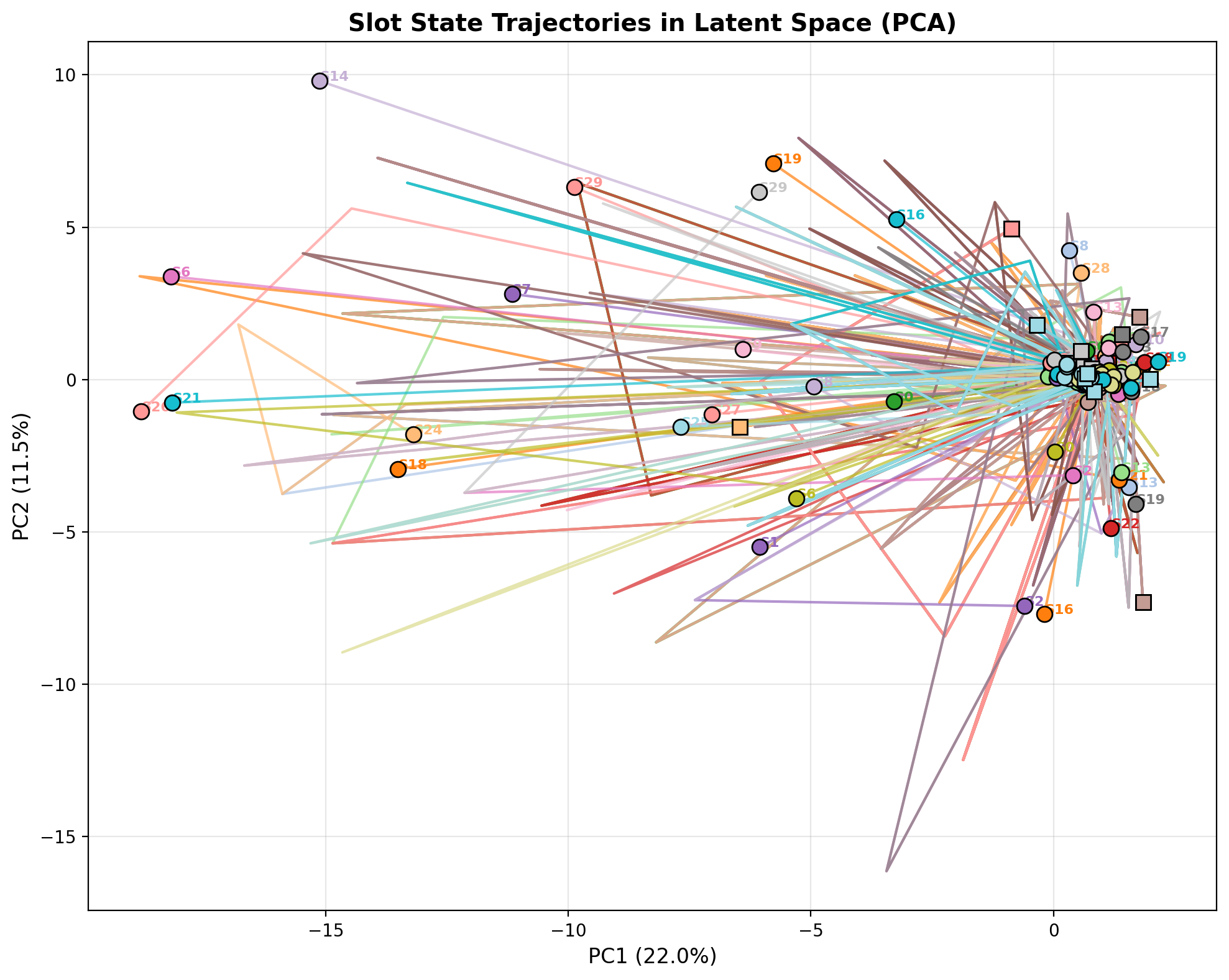}
\caption{Slot state trajectories projected to 2D via PCA. Circles = start, squares = end. Different colors = different slots. Trajectories show structured temporal dynamics with slot-specific evolution paths.}
\label{fig:trajectories}
\end{figure}

Figure~\ref{fig:trajectories} visualizes slot state trajectories projected to 2D via PCA (57\% variance explained in the best run, 33.5\% in the two-stage run). The trajectories show structured temporal evolution: slots follow smooth paths with occasional direction changes at event boundaries. The spread of trajectories indicates that different slots track different aspects of the scene dynamics.

\section{Limitations and Future Work}

We are transparent about where \hclsm{} falls short in its current form. These limitations reflect the bootstrapped nature of this research.

\paragraph{Slot Count.} All 32 slots remain alive throughout training. The existence head does not learn to kill unused slots, meaning a 3-object scene uses 32 slots instead of 3. The spatial decomposition splits each object across $\sim$10 slots rather than assigning one slot per object. Reducing $N_{\text{max}}$ to 8 caused training instability (NaN gradients). Adaptive slot methods~\citep{Fan2024} or codebook-based deduplication~\citep{MetaSlot2025} are promising directions.

\paragraph{Causal Discovery.} The explicit causal adjacency matrix learns no edges (all weights collapse to zero under sparsity regularization). Training the causal graph jointly with dynamics caused NaN at bf16 precision. The GNN edge weights provide an implicit interaction structure, but we have not verified whether these edges correspond to ground-truth causal relationships. A proper evaluation would require environments with known causal structure and intervention-based metrics.

\paragraph{Scale.} We train only the Small configuration (68M) on a single robotics dataset (PushT, 206 episodes). The Base (262M) and Large (3B) configurations have NaN gradient issues at batch size $\geq 4$ that we did not resolve. Multi-GPU training via FSDP failed due to NCCL version incompatibility on our cloud provider. Real production training requires pre-trained ViT initialization (e.g., from V-JEPA 2 weights), larger datasets, and resolved numerical stability.

\paragraph{Seed Sensitivity.} Approximately 40--60\% of training runs diverge to NaN within the first 1000 steps due to seed-dependent gradient overflow in the slot attention GRU at bf16 precision. The surviving runs converge reliably.

\paragraph{Future Directions.}
(1)~Initialize the ViT from V-JEPA 2 checkpoints to provide richer patch features as SBD targets and reduce the perception learning burden.
(2)~Train on diverse, multi-object manipulation data (e.g., ALOHA bimanual tasks with 5+ objects) where the benefits of object-centric decomposition are more pronounced.
(3)~Integrate the CEM/MPPI planners (implemented but not evaluated) with the world model for closed-loop robot control.
(4)~Scale to the Base configuration by resolving the bf16 gradient stability issues, potentially through gradient scaling or fp32 critical-path computation.

\section{Conclusion}

We presented \hclsm{}, an architecture built on a simple premise: world models should mirror the structure of the world they model. Objects are discrete, so representations should be object-centric. Time has multiple scales, so dynamics should be hierarchical. Interactions are causal, so the model should maintain an explicit interaction graph. The two-stage training protocol resolves the tension between these structural commitments and the optimization pressure to collapse into flat distributed codes.

Training on real robot data demonstrates that these components can work together: 0.008 MSE prediction, emerging spatial decomposition, functional event detection, and a 38$\times$ SSM kernel speedup. The architecture is a foundation. Scaling to diverse multi-object scenes with pre-trained vision backbones, resolving the numerical stability barriers at 262M+ parameters, and closing the loop with model-based planning are the paths from foundation to a complete autonomous agent. We release the complete system, warts and all, at \url{https://github.com/rightnow-ai/hclsm}.



\begin{thebibliography}{20}

\bibitem[Jiang(2024)]{Jiang2024}
Jiang, J., Deng, F., Singh, G., Lee, M., and Ahn, S. Slot State Space Models. \emph{arXiv preprint arXiv:2406.12272}, 2024.

\bibitem[Assran(2025)]{Assran2025}
Assran, M. et al. V-JEPA 2: Self-Supervised Video Models Enable Understanding, Prediction and Planning. \emph{Meta AI Research}, 2025.

\bibitem[Bardes(2024)]{Bardes2024}
Bardes, A. et al. V-JEPA: Latent Video Prediction for Visual Representation Learning. \emph{arXiv preprint arXiv:2404.08471}, 2024.

\bibitem[Brohan(2023)]{Brohan2023}
Brohan, A. et al. RT-2: Vision-Language-Action Models Transfer Web Knowledge to Robotic Control. \emph{arXiv preprint arXiv:2307.15818}, 2023.

\bibitem[Sch{\"o}lkopf(2021)]{Scholkopf2021}
Sch{\"o}lkopf, B., Locatello, F., Bauer, S., Ke, N.R., Kalchbrenner, N., Goyal, A., and Bengio, Y. Towards Causal Representation Learning. \emph{Proceedings of the IEEE}, 109(5):612--634, 2021.

\bibitem[Elsayed(2022)]{Elsayed2022}
Elsayed, G. et al. SAVi++: Towards End-to-End Object-Centric Learning from Real-World Videos. \emph{NeurIPS}, 2022.

\bibitem[Fan(2024)]{Fan2024}
Fan, Z. et al. Adaptive Slot Attention: Object Discovery with Dynamic Slot Number. \emph{CVPR}, 2024.

\bibitem[Ghosh(2024)]{Ghosh2024}
Ghosh, D. et al. Octo: An Open-Source Generalist Robot Policy. \emph{arXiv preprint arXiv:2405.12213}, 2024.

\bibitem[Gu(2022)]{Gu2022}
Gu, A. et al. Efficiently Modeling Long Sequences with Structured State Spaces. \emph{ICLR}, 2022.

\bibitem[Gu(2023)]{Gu2023}
Gu, A. and Dao, T. Mamba: Linear-Time Sequence Modeling with Selective State Spaces. \emph{arXiv preprint arXiv:2312.00752}, 2023.

\bibitem[Hafner(2023)]{Hafner2023}
Hafner, D. et al. Mastering Diverse Domains through World Models. \emph{arXiv preprint arXiv:2301.04104}, 2023.

\bibitem[Hu(2023)]{Hu2023}
Hu, A. et al. GAIA-1: A Generative World Model for Autonomous Driving. \emph{arXiv preprint arXiv:2309.17080}, 2023.

\bibitem[Kipf(2022)]{Kipf2022}
Kipf, T. et al. Conditional Object-Centric Learning from Video. \emph{ICLR}, 2022.

\bibitem[LeRobot(2024)]{LeRobot2024}
Cadene, R. et al. LeRobot: State-of-the-art Machine Learning for Real-World Robotics in Pytorch. \emph{GitHub}, 2024. \url{https://github.com/huggingface/lerobot}

\bibitem[Locatello(2020)]{Locatello2020}
Locatello, F. et al. Object-Centric Learning with Slot Attention. \emph{NeurIPS}, 2020.

\bibitem[MetaSlot(2025)]{MetaSlot2025}
Gao, Y. et al. MetaSlot: Break Through the Fixed Number of Slots in Object-Centric Learning. \emph{arXiv preprint arXiv:2505.20772}, 2025.

\bibitem[OpenXEmbodiment(2024)]{OpenXEmbodiment2024}
Open X-Embodiment Collaboration. Open X-Embodiment: Robotic Learning Datasets and RT-X Models. \emph{ICRA}, 2024.

\bibitem[Seitzer(2023)]{Seitzer2023}
Seitzer, M. et al. Bridging the Gap to Real-World Object-Centric Learning. \emph{ICLR}, 2023.

\bibitem[Wu(2023)]{Wu2023}
Wu, Z. et al. SlotFormer: Unsupervised Visual Dynamics Simulation with Object-Centric Models. \emph{ICLR}, 2023.

\bibitem[Yu(2019)]{Yu2019}
Yu, Y. et al. DAG-GNN: DAG Structure Learning with Graph Neural Networks. \emph{ICML}, 2019.

\bibitem[Zheng(2018)]{Zheng2018}
Zheng, X. et al. DAGs with NO TEARS: Continuous Optimization for Structure Learning. \emph{NeurIPS}, 2018.

\end{thebibliography}
\end{document}